\documentclass[10pt,twocolumn,letterpaper]{article}

\usepackage{cvpr}              % To produce the CAMERA-READY version
\usepackage{graphicx}
\usepackage{amsmath}
\usepackage{amssymb}
\usepackage{booktabs}
\usepackage{algorithm}
\usepackage{algpseudocode}

\usepackage[pagebackref,breaklinks,colorlinks]{hyperref}

\usepackage[capitalize]{cleveref}
\crefname{section}{Sec.}{Secs.}
\Crefname{section}{Section}{Sections}
\Crefname{table}{Table}{Tables}
\crefname{table}{Tab.}{Tabs.}

\def\cvprPaperID{3679} % *** Enter the CVPR Paper ID here
\def\confName{CVPR}
\def\confYear{2022}

\begin{document}

\title{Continual Test-Time Domain Adaptation} 

\author{
Qin Wang\textsuperscript{1} \quad  Olga Fink\textsuperscript{1,3\thanks{The corresponding author}} \quad Luc Van Gool\textsuperscript{1,4} \quad Dengxin Dai\textsuperscript{2}\\
\textsuperscript{1}ETH Zurich, Switzerland\space\space\textsuperscript{2}MPI for Informatics, Germany\space\space\textsuperscript{3}EPFL, Switzerland\space\space \textsuperscript{4}KU Lueven, Belgium\\
{\tt\small  \{qin.wang,vangool,dai\}@vision.ee.ethz.ch olga.fink@epfl.ch}
}
\maketitle

%%%%%%%%% ABSTRACT
\begin{abstract}
Test-time domain adaptation aims to adapt a source pre-trained model to a target domain without using any source data. Existing works mainly consider the case where the target domain is static. However, real-world machine perception systems are running in non-stationary and continually changing environments where the target domain distribution can change over time. Existing methods, which are mostly based on self-training and entropy regularization, can suffer from these non-stationary environments. Due to the distribution shift over time in the target domain, pseudo-labels become unreliable. The noisy pseudo-labels can further lead to error accumulation and catastrophic forgetting. To tackle these issues, we propose a continual test-time  adaptation approach~(CoTTA) which comprises two parts. Firstly, we propose to reduce the error accumulation by using weight-averaged and augmentation-averaged predictions which are often more accurate. On the other hand, to avoid catastrophic forgetting, we propose to stochastically restore a small part of the neurons to the source pre-trained weights during each iteration to help preserve source knowledge in the long-term. The proposed method enables the long-term adaptation for all parameters in the network. CoTTA is easy to implement and can be readily incorporated in off-the-shelf pre-trained models. We demonstrate the effectiveness of our approach on four classification tasks and a segmentation task for continual test-time  adaptation, on which we outperform existing  methods. Our code is available at \url{https://qin.ee/cotta}.

\end{abstract}

%%%%%%%%% BODY TEXT
\section{Introduction}
% test-time domain adaptation adapts on test-time
Test-time domain adaptation aims to adapt a source pre-trained model by learning from the unlabeled test~(target) data during inference time. Due to the domain shift between source training data and target test data, an adaptation is necessary to achieve good performance. For example, a semantic segmentation model trained on data from clear weather conditions can suffer significant performance deterioration when tested on snowy night conditions~\cite{SDV21}. Similarly, a pre-trained image classification model can also suffer this phenomenon when tested on corrupted images resulting from sensor degradation.  Due to privacy concerns or legal constraints, the source data is generally considered unavailable during inference time under this setup, making it a more challenging but more realistic problem than unsupervised domain adaptation. In many scenarios, the adaptation also needs to be performed in an online fashion. Therefore, test-time adaptation is critical to the success of real-world machine perception applications under domain shift.
\begin{figure}[t!]
	\centering
	\includegraphics[width=\columnwidth ]{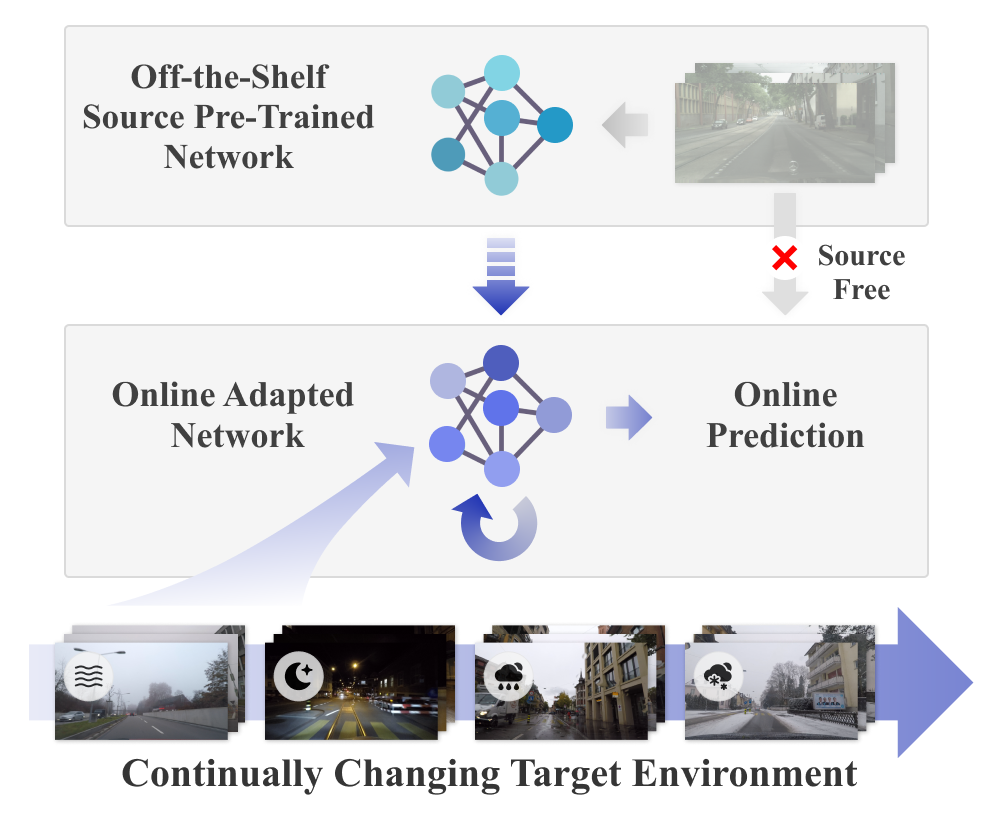}
	\vspace{-1cm}
	\caption{We consider the online continual test-time adaptation scenario. The target data is provided in a sequence and from a continually changing environment. An off-the-shelf source pre-trained network is used to initialize the target network. The model is updated online based on the current target data, and the predictions are given in an online fashion. The adaptation of the target network does not rely on any source data. Existing methods often suffer from error accumulation and forgetting which result in performance deterioration over time. Our method enables long-term test-time adaptation under continually changing environments.}
\vspace{-0.6cm}
	\label{view}
\end{figure}

% Existing works, overlooked continuous problems
%Motivated by the success of self-training and entropy minimization in domain adaptation, 
Existing works on test-time adaptation often tackle the distribution shift between the source domain and a fixed target domain by updating model parameters using pseudo-labels or entropy regularization~\cite{wang2020tent,mummadi2021test}. These self-training methods have been proven to be effective when the test data are drawn from the same stationary domain. However, they can be unstable~\cite{prabhu2021sentry} when the target test data originates from an environment which is continually changing. There are two aspects that contribute to this: Firstly, under the continually changing environment, the pseudo-labels become noisier and mis-calibrated~\cite{guo2017calibration} because of the distribution shift. Therefore, early prediction mistakes are more likely to result in error accumulation~\cite{chen2019progressive}. Secondly, as the model is being continually adapted to new distributions for a long time, knowledge from the source domain is harder to preserve, leading to catastrophic forgetting~\cite{mccloskey1989catastrophic,parisi2019continual,ebrahimi2020adversarial}. 

% To address this problem...
Aiming to tackle these problems under the continually changing environment, this work focuses on the practical problem of online continual test-time  adaptation. As shown in Figure~\ref{view}, the goal is to start from an off-the-shelf source pre-trained model, and continually adapt it to the current test data.  Under this setup, we assume that the target test data is streamed from a continually changing environment. The prediction and updates are performed online, meaning that the model will only have access to the current stream of data without having access to the full test data nor any source data.  The proposed setup is very relevant for real-world machine perception systems. For example, surrounding environments are continually changing for autonomous driving systems~(e.g. weather change from sunny to cloudy then to rainy). They can even change abruptly~(e.g. when a car exits a tunnel and the camera gets suddenly over-exposed). A perception model need to adapt itself and make decisions online under these non-stationary domain shifts.

% Proposed Method
To effectively adapt the pre-trained source model to the continually changing test data, we propose a continual test-time  adaptation approach~(CoTTA) which tackles the two main limitations of existing methods.  The first component of the proposed method aims to alleviate error accumulation. We propose to improve the pseudo-label quality under the self-training framework in two different ways. On the one hand, motivated by the fact that the mean teacher predictions often have a higher quality than the standard model~\cite{tarvainen2017mean}, we use a weight-averaged teacher model to provide more accurate predictions. On the other hand, for test data which suffers larger domain gap, we use the augmentation-averaged predictions to further boost the quality of pseudo-labels. The second component of the proposed method aims to help preserve the source knowledge and avoid forgetting. We propose to stochastically restore a small part of neurons in the network back to the pre-trained source model. By reducing error accumulation and preserving knowledge, CoTTA enables long-term adaptation in a continuously changing environment, and makes it possible to train all parameters of the network. In contrast, previous methods~\cite{wang2020tent,mummadi2021test} can only train batchnorm parameters. 

It is worth pointing out that our approach can be easily implemented. The weight-and-augmentation-averaged strategy and the stochastic restoration can be readily incorporated into any off-the-shelf pre-trained model without the need to re-train it on source data. We demonstrate the effectiveness of our proposed approach on four classification tasks and a segmentation task for continual test-time  adaptation, on which we significantly improve performance over existing methods. Our contributions are summarized blow:
\begin{itemize}
	\item We propose a continual test-time  adaptation approach which can effectively adapt off-the-shelf source pre-trained models to continually changing target data. 
	\item Specifically, we reduce the error accumulation by using weight-averaged and augmentation-averaged pseudo-labels that are more accurate. 
	\item The long-term forgetting effect is alleviated by explicitly preserving the knowledge from the source model.
	\item The proposed approach significantly improves the continual test-time  adaptation performance on both classification and segmentation benchmarks.
\end{itemize}

\section{Related Work}

\subsection{Domain Adaptation}
Unsupervised domain adaptation~(UDA)~\cite{pan2011domain, patel2015visual} aims to improve the target model performance in the presence of a domain shift between the labeled source domain and unlabeled target domain. During training, UDA methods often align the feature distributions between the two domains using discrepancy losses~\cite{long2015learning} or adversarial training~\cite{ganin2014unsupervised,Tsai_adaptseg_2018}. Alternatively, the alignment can also be done in the input space~\cite{hoffman2018cycada, yang2020fda}. In recent years, self-training has also shown promising results by  iteratively using gradually-improving target pseudo-labels to train the network~\cite{zou2018domain,Lian_2019_ICCV,wang2021domain,hoyer2021daformer}.
\begin{table*}[hbt!]
\centering
\caption{The difference between our proposed continual test-time  adaptation and related adaptation settings.}\label{tab:setup}
\vspace{-0.2cm}
\scalebox{0.88}{
\tabcolsep12pt

\begin{tabular}{l|cc|cc}
\hline
                                                     & \multicolumn{2}{c|}{Data} & \multicolumn{2}{c}{Learning}  \\ \cline{2-5} 
Setting                                              & Source  & Target          & Train stage      & Test stage \\ \hline
standard domain adaptation                                    & Yes     & stationary      & Yes              & No         \\
standard test-time training~\cite{sun2020test}                                   & Yes     & stationary      & Yes (aux task)   & Yes        \\
fully test-time adaptation~\cite{wang2020tent}                           & No      & stationary      & No (pre-trained) & Yes        \\ \hline
\multicolumn{1}{l|}{continual test-time adaptation} & No      & continually changing  & No (pre-trained) & Yes        \\ \hline
\end{tabular}
}\vspace{-0.5cm}
\end{table*}
\subsection{Test-time Adaptation}
Test-time adaptation is also referred to as source-free domain adaptation in some references~\cite{kundu2020universal,yang2021generalized}. Unlike domain adaptation which requires access to both source and target data for adaptation, test-time adaptation methods do not require any data from the source domain for adaptation. Some existing works~\cite{li2020model,yeh2021sofa,kurmi2021domain} utilize generative models to support the feature alignment in absence of source data.

Another popular direction is to finetune the source model without explicitly conducting domain alignment. Test entropy minimization~(TENT)~\cite{wang2020tent} takes a pre-trained model and adapts to the test data by updating the trainable parameters in Batchnorm layers using entropy minimization. Source hypothesis transfer~(SHOT)~\cite{liang2020we} utilizes both entropy minimization and a diversity regularizer for the adaptation. SHOT requires using source data to train a specialized source model using the label-smoothing technique with the weight normalization layer. Thus, it cannot support the use an arbitrary pre-trained model. %\cite{valvano2021stop} re-uses the discriminator in inference time to strengthen the performance for medical segmentation.
\cite{mummadi2021test} proposes to apply a diversity regularizer combined with an input transformation module to further improve the performance. \cite{karani2021test} uses a separate normalization convolutional network to normalize test images from new domains.  \cite{iwasawa2021test} only updates the final classification layer during inference time using pseudo-prototypes.  \cite{zhou2021training} analyzes the problem in a Bayesian perspective and  proposes a regularized entropy minimization procedure at test-time adaptation, which requires approximating density during training time. Updating the statistics in the Batch Normalization layer using the target data is a different path which also shows promising results~\cite{li2016revisiting,hu2021mixnorm,you2021test}. While most existing works focus on image classification, \cite{liu2021source,kundu2021generalize,hu2021fully} extend test-time adaptation to semantic segmentation. Standard test-time adaptation considers the offline scenario where access to the full set of test data is provided for the training. This is often unrealistic for online machine perception applications. Most existing works~(except TENT variants~\cite{wang2021target}) also require the re-training of the source model to support the test-time adaptation. Therefore, they cannot directly use off-the-shelf pre-trained model from the source domain.

\subsection{Continuous Domain Adaptation}
Unlike standard domain adaptation which assumes a specific target domain, continuous domain adaptation considers the adaptation problem with continually changing target data. Continuous Manifold Adaptation~(CMA)~\cite{hoffman2014continuous} is an early work which considers adaptation to evolving domains. Incremental adversarial domain adaptation~(IADA)~\cite{wulfmeier2018incremental} adapts to continually changing domains by adversarially aligning source and target features. \cite{volpi2021continual} aims to continually adapt the unseen visual domain while alleviate the forgetting on the seen domain without retaining the source training data. \cite{bobu2018adapting} aims to adapt to gradually changing domains by making use of the assumption of continuity between gradually varying domains. Existing continuous domain adaptation methods need to have access to data from both the source and target domains in order to align the distributions. 

The main focus of this paper is continual test-time adaptation, which additionally considers the adaptation at test-time without accessing the source data.  While this is a realistic scenario for machine perception systems in the real world, there are very limited number of approaches which are applicable to such scenarios.  In theory, the online version of TENT~\cite{wang2020tent} could adapt under this setup by continually updating the BN parameters using the entropy loss. However, it can suffer from  error accumulation because of mis-calibrated predictions. Test-time training~(TTT)~\cite{sun2020test} could also continually update the feature extractor using supervision from the rotation prediction auxiliary task. However, it requires re-training of the source model using the source data to learn the auxiliary task. Therefore, it cannot be considered as source-free for the full pipeline and does not support off-the-shelf source pre-trained models. %Several papers on \cite{cai2021online} investigates the online continual learning under domain shifts, where new incoming data is first tested and later added to the training set. However, this requires the knowledge of ground truth labels for the newly added data. The truly online continual test-time  adaptation scenario for arbitrary pre-trained model is largely unexplored.

\subsection{Continual Learning}
Continual learning~\cite{delange2021continual} and lifelong learning~\cite{parisi2019continual} are closely related to the continuous adaptation problems as a potential cure to the catastrophic forgetting. Continual learning methods can often be categorized into replay-based~\cite{rebuffi2017icarl} and regularization-based~\cite{silver2002task,zenke2017continual} methods. The latter can further be divided into data-focused methods, such as learning without forgetting~(LwF)~\cite{li2017learning}, and prior-focused methods, such as elastic weight consolidation~(EWC)~\cite{kirkpatrick2017overcoming}. Ideas from continual learning are adopted for continuous domain adaptation approaches~\cite{bobu2018adapting,lao2020continuous}.

\subsection{Domain Generalization}
This work is also related to  domain generalization~\cite{muandet2013domain} in a broad sense, because of the shared goal of improving performance on potentially changing target domains. A number of works have also shown that data augmentation~\cite{shorten2019survey} during training~\cite{yin2019fourier,hendrycks2020augmix,hendrycks2021many,li2021feature} and during testing~\cite{ashukha2020pitfalls,zhang2021memo,molchanov2020greedy} can improve model robustness and generalizability. Domain randomization is one of the most popular methods which improves the model generalizability by learning from different synthesis parameters of simulation environments~\cite{tobin2017domain, tremblay2018training}.  Unlike domain generalization methods which mostly aim to train a more generalizable neural network  from the source domain, this work focuses on improving the performance of existing pre-trained neural networks during test-time by using the unlabeled online data from the continually changing target domain.

\section{Continual Test-Time Domain Adaptation} 
\subsection{Problem Definition}
Given an existing pre-trained model $f_{\theta_0}(x)$ with parameters $\theta$ trained on the source data $(\mathcal{X}^S, \mathcal{Y}^S)$, we aim at improving the performance of this existing model during inference time for a continually changing target domain in an online fashion without having access to any source data. Unlabeled target domain data~$\mathcal{X}^T$ is provided sequentially and the model only have access to the data of the current time step. At time step $t$, target data $x^T_t$ is provided as input and the model $f_{\theta_t}$ needs to make the prediction $f_{\theta_t}(x^T_t)$ and adapts itself accordingly for future inputs $\theta_{t}\xrightarrow{}\theta_{t+1}$. The data distribution of $x^T_t$ is continually changing. The model is evaluated based on the online predictions.

This setup is largely motivated by the need of machine perception applications in continually changing environments. For example, the surrounding environment is continually changing for  autonomous driving cars  because of location, weather, and time. Perception decisions need to be made online and models need to be adapted.

We list the main differences between our online continual test-time adaptation setup with existing adaptation setups in Table~\ref{tab:setup}. Compared to previous setups which focus on a fixed target domain, we consider the long-term adaptation on continually changing target environments. 
\begin{figure}[t!]
	\centering
	\includegraphics[width=\columnwidth ]{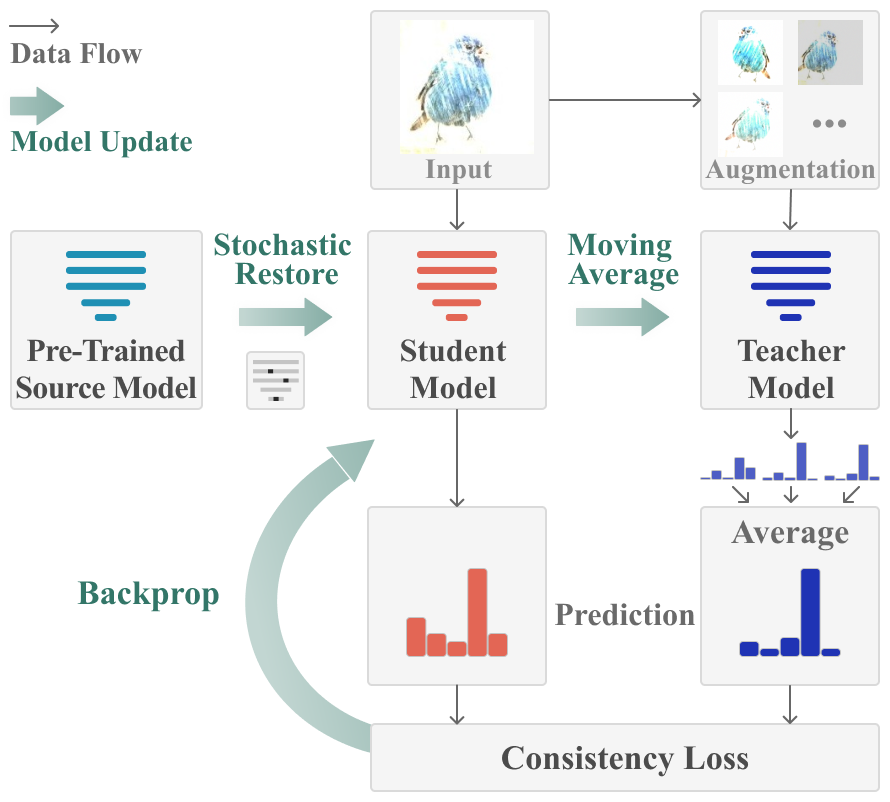}
	\vspace{-0.8cm}
	\caption{An overview of the proposed continual test-time adaptation~(CoTTA) approach. CoTTA adapts from an off-the-shelf source pre-trained network. Error accumulation is mitigated by using a teacher model to provide weight-averaged pseudo-labels and using multiple augmentations to average the predictions. Knowledge from the source data is preserved by stochastically restoring a small number of elements of trainable weights.}
\vspace{-0.6cm}
	\label{fig2}
\end{figure}

\subsection{Methodology}
We propose an adaptation method for the online continual test-time adaptation setup. The proposed method takes an off-the-shelf source pre-trained model and adapts it to the continually changing target data in an online fashion. Motivated by the fact that error accumulation is one of the key bottlenecks in the self-training framework, we propose to use weight-and-augmentation-averaged pseudo-labels to reduce error accumulation. In addition, to help reduce forgetting in continual adaptation, we propose to explicitly preserve information from the source model. An overview of the proposed method is presented in Figure~\ref{fig2}.

% motivation
\paragraph{Source Model}
Existing works on test-time adaptation often require special treatment in the training process of the source model to improve domain generalization ability and to facilitate the adaptation. For example, %SHOT~\cite{liang2020we} trains the source model using the label-smoothing technique with an additional weight normalization layer.
during source training, TTT~\cite{sun2020test} has an additional auxiliary rotation prediction branch to train to facilitate the target adaptation supervision. This requires a retraining on the source data, and makes it impossible to reuse existing pre-trained models. In our proposed test-time adaptation method, we lift this burden and do not require a modification of the architecture or an additional source training process. %The adaptation is purely performed at the test-time. 
Therefore, any existing pre-trained models can be used without retraining on the source. We will show in the experiments that our method can work on a wide range of pre-trained networks including ResNet variants and Transformer-based architectures. 
\paragraph{Weight-Averaged Pseudo-Labels}
Given target data $x^T_t$ and the model $f_{\theta_t}$, the common test-time objective under the self-training framework is to minimize the cross-entropy consistency between the prediction $\hat{y}^T_t=f_{\theta_t}(x^T_t)$ and a pseudo-label. For example, directly using the model prediction itself as the pseudo-label leads to the training objective of TENT~\cite{wang2020tent}~(i.e. entropy minimization). While this works for a stationary target domain, the quality of pseudo-labels can drop significantly for continually changing target data because of the distribution shift.

Motivated by the observation that weight-averaged models over training steps often provide a more accurate model than the final model~\cite{polyak1992acceleration, tarvainen2017mean}, we use a weight-averaged teacher model $f_{\theta'}$ to generate the pseudo-labels. At time-step $t=0$, the teacher network is initialized to be the same as the source pre-trained network. At time-step $t$, the pseudo-label is first generated by the teacher $\hat{y'}^T_t=f_{\theta'_t}(x^T_t)$. The student $f_{\theta_t}$ is then updated by the cross-entropy loss between the student and teacher predictions:

\begin{equation}
\mathcal{L}_{\theta_t}(x^T_t) = -\sum_c \hat{y'}^T_{tc} \log  \hat{y}^T_{tc}, \label{eq:weight_consistency}
\end{equation}
where $\hat{y'}^T_{tc}$ is the probability of class c in the teacher model's soft pseudo-label prediction, and $\hat{y}^T_{tc}$ is the prediction from the main model~(student). The loss enforces a consistency between the teacher  and  student predictions.

After the update of the student model $\theta_t \xrightarrow{} \theta_{t+1}$ using Equation~\ref{eq:weight_consistency}, we update the weights of the teacher model by exponential moving average using the student weights: 
\begin{equation}
\theta'_{t+1} =  \alpha \theta'_{t} + (1-\alpha) \theta_{t+1}, \label{ema}
\end{equation}
where $\alpha$ is a smoothing factor. Our final prediction for the input data $x^T_t$ is the class with the highest probability in $\hat{y'}^T_t$. %Both the student and teacher models are are continuously optimized based on the current data.

The benefits of the weight-averaged consistency are two-fold. On the one hand, by using the often more accurate~\cite{polyak1992acceleration} weight-averaged prediction as the pseudo-label target, our model suffers less from the error accumulation during the continual adaptation.  
On the other hand, the mean teacher prediction $\hat{y'}^T_t$ encodes the information from models in past iterations and is, therefore, less likely to suffer from catastrophic forgetting in long-term continual adaptation and improve the generalization capability to new unseen domains. This is inspired by the mean teacher method proposed in~\cite{tarvainen2017mean} in semi-supervised learning. 
\paragraph{Augmentation-Averaged Pseudo-Labels}
Data augmentation during training time~\cite{shorten2019survey} has been widely applied to improve model performance. Different augmentation policies are often manually designed~\cite{krizhevsky2012imagenet} or searched~\cite{cubuk2019autoaugment} for different datasets. While test-time augmentation has also been proven to be able to improve robustness~\cite{sun2020test, cohen2021katana}, the augmentation policies are generally determined and fixed for a specific dataset without considering the distribution change during inference time. Under a continually changing environment, test distributions can change dramatically, which may make the augmentation policy invalid. 
Here, we take the test-time domain shift into account and approximate the domain difference by prediction confidence. The augmentation is only applied when the domain difference is large, to reduce error accumulation. 
%Therefore, in addition to using the weight-averaged teacher model to improve pseudo-label quality, we additionally make use of data augmentation to reduce error accumulation.  Specifically, we apply N augmentations on the target samples with a large domain gap:  
\begin{align}
   %\hat{y'}^T_t&= f_{\theta'_t}(x^T_t),\\
   \tilde{y'}^T_t&= \frac{1}{N}\sum_{i=0}^{N-1}f_{\theta'_t}(\text{aug}_i(x^T_t)),\label{eq:aug}\\
   {y'}^T_t &= \begin{cases}
      \hat{y'}^T_t, & \text{if} \text{ conf}(f_{\theta_0}(x^T_t)) \geq p_{th}\\ %\text{conf}^S-\delta \\
      \tilde{y'}^T_t, & \text{otherwise},\label{eq:if}
    \end{cases} 
    \end{align}
where  $\tilde{y'}^T_t$ is the augmentation-averaged prediction from the teacher model, $\hat{y'}^T_t$ is the direct prediction from the teacher model, $\text{ conf}(f_{\theta_0}(x^T_t))$ is the source pre-trained model's prediction confidence on the current input  $x_t^T$, and $p_{th}$ is a confidence threshold. By calculating the prediction confidence on the current input $x_t^T$ using the pre-trained model $f_{\theta_0}$ in Equation~\ref{eq:if}, we attempt to approximate the domain difference between the source and the current domain. We hypothesize that a lower confidence indicates a larger domain gap and a relatively high confidence level indicates a smaller domain gap. Therefore, when the confidence is high and larger than the threshold, we directly use $\hat{y'}^T_t$ as our pseudo-label without using any augmentation. When the confidence is low, we apply additionally N random augmentations to further improve the pseudo-label quality. The filtering is critical as we observe that random augmentations on confident samples with small domain gaps can sometimes decrease model performance. We provide detailed discussion on this observation in the supplementary. In summary, we use the confidence to approximate the domain difference and determine when to apply the augmentations.

The student is updated by the refined pseudo-label:
\begin{equation}
\mathcal{L}_{\theta_t}(x^T_t) = -\sum_c {y'}^T_{tc} \log  \hat{y}^T_{tc}, \label{eq:weight_consistency2}
\end{equation}%$\text{conf}^S$ is the pre-trained model's average confidence on the source data, and $\delta$ is small tolerance factor. T
\begin{algorithm}[t]
\caption{The proposed continual test-time adaptation}
\label{algo}
% Initialize the CDF $\bar{F}$ with an identity map.
\textbf{Initialization: } A source pre-trained model $f_{\theta_0}(x)$, teacher model $f_{\theta'_0}(x)$ initialized from $f_{\theta_0}(x)$. \\
\textbf{Input: } For each time step $t$, current stream of data $x_t$. 
\begin{algorithmic}[1]
\State Augment $x_t$ and get weight and augmentation-averaged pseudo-labels from the teacher $f_{\theta'_t}$ by Equation~\ref{eq:if}.
\State Update student $f_{{\theta}_t}$ by consistency loss in Equation~\ref{eq:weight_consistency2}.
\State Update teacher $f_{\theta'_t}$ by moving average in Equation~\ref{ema}.
\State Stochastically restore student $f_{\theta_t}$ by Equation~\ref{restore}.

\end{algorithmic}
\textbf{Output:} Prediction $f_{\theta'_t}(x_t)$; Updated student model $f_{\theta_{t+1}}(x)$; Updated teacher model $f_{\theta'_{t+1}}(x)$.
\end{algorithm} \vspace{-5mm}
\paragraph{Stochastic Restoration}
While more accurate pseudo-labels can mitigate error accumulation, continual adaptation by self-training for a long time inevitably introduces errors and leads to forgetting. This issue can be especially relevant if we encounter strong domain shifts within a sequence of data, because the strong distribution shift  leads to mis-calibrated and even wrong predictions. Self-training in this case may only lead to reinforcing wrong predictions. What's worse is that after encountering hard examples, the model may not be able to recover because of the continual adaptation, even when the new data are not severely shifted. 

To further tackle the problem of catastrophic forgetting, we propose a stochastic restoration method which explicitly restores the knowledge from the  source pre-trained model.

Consider a convolution layer within the student model $f_\theta$ after gradient update based on Equation~\ref{eq:weight_consistency} at time step $t$:
\begin{equation}
    x_{l+1} = W_{t+1} * x_l, 
\end{equation}
where $*$ denotes the convolution operation, $x_l$ and $x_{l+1}$ denote the input and output to this layer, $W_{t+1}$ denotes the trainable convolution filters. The proposed stochastic restoration method additionally updates the weight $W$ by:
\begin{align}
    M &\sim \text{Bernoulli}(p),\\
    W_{t+1} &= M \odot W_{0}  + (1-M) \odot W_{t+1} , \label{restore}
\end{align}
where $\odot$ denotes the element-wise multiplication. $p$ is a small restore probability, and $M$ is a mask tensor of the same shape as $W_{t+1}$. The mask tensor decides which element within $W_{t+1}$ to restore back to the source weight $W_0$. 

The stochastic restoration can also be seen as a special form of Dropout. By stochastically restoring a small number of tensor elements in the trainable weights to the initial weight, the network avoids drifting too far away from the initial source model and therefore, avoids catastrophic forgetting. In addition, by preserving the information from the source model, we are able to train all trainable parameters without suffering from model collapse. This brings more capacity for the adaptation and is another major difference compared to entropy minimization methods~\cite{wang2020tent, mummadi2021test} which only train the BN parameters for test-time adaptation. 

As shown in Algorithm~\ref{algo}, combining the refined pseudo-labels  with stochastic restoration leads to our online continual test-time  adaptation~(CoTTA) method.

\section{Experiments}

We evaluate our proposed method on five continual test-time adaptation benchmark tasks: CIFAR10-to-CIFAR10C (standard and gradual), CIFAR100-to-CIFAR100C, and ImageNet-to-ImageNet-C for image classification, as well as Cityscapses-to-ACDC for semantic segmentation. %A gradually changing test-time adaptation classification task is also included for CIFAR10C.

\subsection{Datasets and tasks}
\paragraph{CIFAR10C, CIFAR100C, and ImageNet-C} were originally created to benchmark robustness of classification networks~\cite{hendrycks2019robustness}. Each dataset contains 15 types of corruptions with 5 levels of severity.  The corruptions were applied on images from the test set of the clean CIFAR10 or CIFAR100 dataset~\cite{krizhevsky2009learning}. There are 10,000 images for each corruption type for both CIFAR10C and CIFAR100C datasets. 

For our online continual test-time adaptation task, a network pre-trained on the clean training set of CIFAR10 or CIFAR100 dataset is used. During test time, the corrupted images are provided in an online fashion to the network. Unlike previous methods which evaluate the test-time adaptation performance from the clean images pre-trained model to each corruption type individually, we continually adapt the source pre-trained model to each corruption type sequentially. %Specifically, we follow the default corruption order: Gaussian noise, shot noise, impulse noise, defocus blur, glass blur, motion blur, zoom blur,	snow,	frost, fog,	brightness,	contrast,	elastic transformation,	pixelation, and	jpeg compression. %The corrupted data are streamed online to the model in this order, and the model is updated for each streamed iteration. After the streaming is finished for one corruption type, we continuously adapt and update our model for the streams from the next corruption type. 
We evaluate all models under the largest corruption severity level 5. The evaluation is based on the online prediction results immediately after the encounter of the data. Both the CIFAR10  and CIFAR100  experiments follow this online continual test-time adaptation scheme. 

For CIFAR10-to-CIFAR10C, we follow the official public implementation from TENT~\cite{wang2020tent} for the CIFAR10 experiments. The same pre-trained model is adopted, which is a WideResNet-28~\cite{zagoruyko2016wide} model from the RobustBench benchmark~\cite{croce2021robustbench}. We update the model for one step at each iteration (i.e. one gradient step per test
point). We use the same Adam optimizer with a learning rate of 1e-3 as the official implementation. Following~\cite{cohen2021katana}, we use the same random augmentation composition including color jitter, random affine, gaussian blur, random horizonal flip, and gaussian noise. We use 32 augmentations for our experiments. We discuss the choice of the augmentation threshold $p_{th}$ in our supplementary material. Unlike TENT models which only update the BN scale and shift weights, we update all trainable parameters in the experiments.  We use a restoration probability of $p=0.01$ for all our experiments. 

For CIFAR100-to-CIFAR100C experiments, we adopt the pre-trained ResNeXt-29~\cite{xie2017aggregated} model from ~\cite{hendrycks2020augmix}, which is used as one of the default architectures for CIFAR100 in the RobustBench benchmark~\cite{croce2021robustbench}.  The same hyperparameters are used as in the CIFAR10 experiments. The ImageNet-to-ImageNet-C~\cite{hendrycks2019robustness} experiments use the \textit{standard pre-trained resnet50} model in RobustBench~\cite{croce2021robustbench}. ImageNet-C experiments are evaluated under ten diverse corruption orders.

\paragraph{Cityscapes-to-ACDC} is a continual semantic segmentation task we designed to mimic continual distribution shifts in the real world. The source model is an off-the-shelf pre-trained segmentation model trained on the Cityscapes dataset~\cite{Cordts2016Cityscapes}. The target domain contains images from various scenarios from the Adverse Conditions Dataset~(ACDC)~\cite{SDV21}. The ACDC dataset shares the same  semantic classes with Cityscapes and is collected in four different adverse visual conditions: Fog, Night, Rain, and Snow. We evaluate our continual test-time adaptation following the same default order. We use 400 unlabeled images from each adverse condition for the adaptation. To mimic the scenario in real life where similar environments might be revisited, and to evaluate the forgetting effect of our methods, we repeat the same sequence group~(of the four conditions) 10 times (i.e. in total 40: Fog$\xrightarrow[]{}$Night$\xrightarrow[]{}$Rain$\xrightarrow[]{}$Snow$\xrightarrow[]{}$Fog...). This also provides an evaluation of the adaptation performance in the long term.%More specifically, at each iteration, the model is streamed with one image from the current adverse condition. After we finish the adaptation for the last image from the last adverse condition~(snow), we continually repeat the stream and feed the first image from the first condition fog. In total, we visit each adverse condition for ten times. This repetition is designed to evaluate the long-term performance and level of forgetting of the adaptation methods. It mimics the scenario in real life where similar environment might be revisited for perception machines. We use 400 images from each adverse conditions for the adaptation. No ground truth labels are used for our adaptation.

For the implementation details, we adopt a transformer-based architecture, Segformer~\cite{xie2021segformer}, for our Cityscapse-to-ACDC experiments. We use the publicly-available pre-trained Segformer-B5 trained on Cityscapes as our off-the-shelf source model. %It is worth mentioning that the Segformer models use mostly LayerNorm~\cite{ba2016layer} in the main backbone. 
For the baseline comparison method, $TENT$ optimizes the parameters in the normalization layers. For the proposed CoTTA model, all trainable layers are updated without the need to choose specific layers. Images from ACDC have a resolution of 1920x1080. We use down-sampled resolutions of 960x540 as inputs to the network and the predictions are evaluated under the original resolution.  Adam optimizer is used with the learning rate 8 times smaller than the default one for Segformer, because we use batch size 1 instead of 8~(default for source training) in our online continual test-time adaptation experiments. We use the multi-scaling input with flip as the augmentation method for the proposed method to generate augmentation-weighted pseudo-label~(as in Equation~\ref{eq:aug}). Following the default practice designed for Cityscapes in MMSeg~\cite{mmseg2020}, we use the scale factors of [0.5, 0.75, 1.0, 1.25, 1.5, 1.75, 2.0]. %Together with the flip operation, 14 copies of augmented images are forwarded in the network for each input image for the augmentation-weighted pseudo-label. 
\begin{table*}[ht!]
\centering
\caption{Classification error rate~(\%) for the standard CIFAR10-to-CIFAR10C online continual test-time adaptation task. Tesults are evaluated on WideResNet-28 with the largest corruption severity level 5. * denotes the requirement on additional domain information. }\label{tab:cifar10}\vspace{-0.2cm}
\vspace{-2.5mm}
\scalebox{0.76}{
\tabcolsep3pt
\begin{tabular}{l|   c|c|c|               ccccccccccccccc|c}
\multicolumn{4}{l}{}& \multicolumn{15}{l}{ $t\xrightarrow{\hspace*{12.5cm}}$}& \\ \hline
Method   & \rotatebox[origin=c]{70}{\parbox{1.4cm}{\centering Weight-\\ avg.}} & \rotatebox[origin=c]{70}{\parbox{1.4cm}{\centering Aug-\\ avg.}}                      & \rotatebox[origin=c]{70}{\parbox{1.4cm}{\centering Stochastic\\ Restore}}  & \rotatebox[origin=c]{70}{Gaussian} & \rotatebox[origin=c]{70}{shot} & \rotatebox[origin=c]{70}{impulse} & \rotatebox[origin=c]{70}{defocus} & \rotatebox[origin=c]{70}{glass} & \rotatebox[origin=c]{70}{motion} & \rotatebox[origin=c]{70}{zoom} & \rotatebox[origin=c]{70}{snow} & \rotatebox[origin=c]{70}{frost} & \rotatebox[origin=c]{70}{fog}  & \rotatebox[origin=c]{70}{brightness} & \rotatebox[origin=c]{70}{contrast} & \rotatebox[origin=c]{70}{elastic\_trans} & \rotatebox[origin=c]{70}{pixelate} & \rotatebox[origin=c]{70}{jpeg} & Mean \\ \hline
Source   & & &                         & 72.3     & 65.7 & 72.9    & 46.9    & 54.3  & 34.8   & 42.0 & 25.1 & 41.3   & 26.0 & 9.3        & 46.7     & 26.6           & 58.5     & 30.3 & 43.5 \\
BN Stats Adapt   & & &                      & 28.1     & 26.1 & 36.3    & 12.8    & 35.3  & 14.2   & 12.1 & 17.3 & 17.4   & 15.3 & 8.4        & 12.6     & 23.8           & 19.7     & 27.3 & 20.4 \\
Pseudo-label  & & & &26.7&	22.1&	32.0&	13.8&	32.2&	15.3&	12.7&	17.3&	17.3&	16.5&	10.1&	13.4&	22.4&	18.9&	25.9&	19.8 \\
TENT-online*~\cite{wang2020tent}      & & &                         & 24.8     & 23.5 & 33.0    & 12.0    & 31.8  & 13.7   & 10.8 & 15.9 & 16.2   & 13.7 & 7.9        & 12.1     & 22.0           & 17.3     & 24.2 & 18.6 \\
TENT-continual~\cite{wang2020tent}     & & &        & 24.8 &	\textbf{20.6} &	28.6 &	14.4 &	31.1 &	16.5 &	14.1 &	19.1 &	18.6 &	18.6 &	12.2 &	20.3 &	25.7 &	20.8 &	24.9 &	20.7\\%24.8     & 20.5 & 28.6    & 15.1    & 31.3  & 15.8   & 14.6 & 18.6 & 19.6   & 20.9 & 14.7       & 19.6     & 28.1           & 23.9     & 29.1 & 21.7 \\
%TENT+lwf (continuous) & 24.8& 	20.7& 	29.2& 	13.8& 	31.0& 	14.6& 	12.2& 	16.5& 	15.3& 	14.1& 	8.9& 	12.9& 	21.3& 	16.2& 	21.9&  18.2\\
\hline

%Weight                 & 27.3     & 23.5 & 32.4    & 11.9    & 30.7  & 12.3   & 10.6 & 15.2 & 14.5   & 12.5 & 7.7        & 10.9     & 18.3           & 13.8     & 19.6 & 17.4 \\
CoTTA (Ours) & \checkmark & & &27.2&22.8&30.8& 12.1& 30.1 & 13.9 & 11.9 & 17.2 & 16.0 &14.3 &9.4& 13.1 & 19.9 & 15.4 &19.9 & 18.3 \\
%Aug-averaged (ours) & & \checkmark & & 25.0&	23.3&	28.3&	12.8&	31.4&	14.2&	12.1&	17.3&	17.3&	15.3&	8.4&	12.6&	23.4&	19.5&	23.4&	18.9\\
CoTTA (Ours) &\checkmark&\checkmark& &  24.5 & 21.0 & \textbf{26.0} & 12.3 & 27.9 & 13.9 & 12.0 & 16.6 & 15.9 & 14.7 & 9.4 & 13.6 & 19.8 & 14.7 & 18.7 & 17.4               \\ 
CoTTA~(Ours)  &\checkmark &\checkmark &\checkmark & \textbf{24.3}     & 21.3 & 26.6    & \textbf{11.6}    & \textbf{27.6}  & \textbf{12.2}   & \textbf{10.3} & \textbf{14.8} & \textbf{14.1}   & \textbf{12.4} & \textbf{7.5}        & \textbf{10.6}    & \textbf{18.3}           & \textbf{13.4}     & \textbf{17.3} & \textbf{16.2}~(0.1)\\\hline
\end{tabular}} \vspace{-0.3cm}
\end{table*}

% \begin{tabular}{l|c|c|c|c}
% \hline
% Method                    & \rotatebox[origin=c]{90}{\parbox{1.4cm}{\centering Weight-\\ avg.}} & \rotatebox[origin=c]{90}{\parbox{1.4cm}{\centering Stochastic\\ Restore}} & \rotatebox[origin=c]{90}{\parbox{1.4cm}{\centering Aug-\\ avg.}} & \multicolumn{1}{l}{error} \\ \hline
% Source                    &                             &                              &                          & 43.5                           \\
% BN Stat Adapt             &                             &                              &                          & 20.4                           \\ 
% TENT-Continual             &                             &                              &                          & 21.7                         \\ \hline
%  & \checkmark                          &                              &                          & 18.3                             \\
%                           & \checkmark                           & \checkmark                          &                          & 17.4                           \\
%     Proposed                      & \checkmark                           & \checkmark                           & \checkmark                      & 16.5                           \\ \hline
% \end{tabular}
% }
% \end{table}

\begin{table}[t]
\caption{Gradually changing CIFAR10-to-CIFAR10C results. The severity level changes gradually between the lowest and the highest. The corruption type changes when the severity is the lowest. Results are the mean over ten diverse corruption type sequences.}\label{gradual}
\small
\vspace{-2mm}
\centering
\scalebox{0.84}{
\tabcolsep5pt
\begin{tabular}{c|c|c|c|c}
\hline
Avg. Error (\%) & Source & BN Adapt& TENT-continual~\cite{wang2020tent}      & CoTTA           \\\hline
CIFAR10C   & 24.8   &   13.7   & 30.7  & 10.4 $\pm$ 0.2   \\\hline

\hline
\end{tabular}}
\vspace{-7mm}
\end{table}

\subsection{Experiments on CIFAR10-to-CIFAR10C}
We first evaluate the effectiveness of the proposed model on the CIFAR10-to-CIFAR10C task. We compare our method to the source-only baseline and four popular methods. %We report the error rate of the online predictions for each corruption type as well as the average performance. As mentioned earlier, the adaptation is conducted continuously for a sequence of 15 types of corruptions~(at severity level 5) in an online fashion. 
As shown in Table~\ref{tab:cifar10}, directly using the pre-trained model without adaptation yields a high average error rate of 43.5\%, indicating that an adaptation is necessary. The \textit{BN Stats Adapt} method keeps the network weights and uses the Batch Normalization statistics from the input data of the current iteration for the prediction~\cite{li2016revisiting,schneider2020improving}. The approach is simple and fully online, and significantly improves the performance over the source-only baseline. Using hard pseudo-labels~\cite{lee2013pseudo} to update the BN-trainable parameters can reduce the error rate to 19.8\%.  If the \textit{TENT-online}~\cite{wang2020tent} method has access to the additional domain information and resets itself to the initial pre-trained model whenever it encounters a new domain, the performance can be further improved to 18.6\%. However, such information is usually unavailable in real applications. Without having access to this additional information, the \textit{TENT-continual} method does not yield any improvement over the \textit{BN Stats Adapt} method. It is worth mentioning that in earlier stages of the adaptation, \textit{TENT-continual} outperforms the \textit{BN Stats Adapt}. However, the model quickly deteriorates after observing three types of corruptions. This indicates that \textit{TENT} based methods can be unstable under continual adaptation in the long term because of error accumulation. Our proposed method can continuously outperform all the above methods by using the weight-and-augmentation-averaged consistency. The error rate is significantly reduced to 16.2\%. In addition, it does not suffer from performance deterioration in the long term because of our stochastic restore approach.

\begin{table*}[h]
\centering
\caption{Classification error rate~(\%) for the standard CIFAR100-to-CIFAR100C online continual test-time adaptation task. All results are evaluated on the ResNeXt-29 architecture with the largest corruption severity level 5. }\label{tab:cifar100}
\vspace{-3mm}
\scalebox{0.82}{
\tabcolsep4pt
\begin{tabular}{l|lllllllllllllll|c}\hline
Time & \multicolumn{15}{l|}{$t\xrightarrow{\hspace*{13.5cm}}$}& \\ \hline
Method                          & \rotatebox[origin=c]{70}{Gaussian} & \rotatebox[origin=c]{70}{shot} & \rotatebox[origin=c]{70}{impulse} & \rotatebox[origin=c]{70}{defocus} & \rotatebox[origin=c]{70}{glass} & \rotatebox[origin=c]{70}{motion} & \rotatebox[origin=c]{70}{zoom} & \rotatebox[origin=c]{70}{snow} & \rotatebox[origin=c]{70}{frost} & \rotatebox[origin=c]{70}{fog}  & \rotatebox[origin=c]{70}{brightness} & \rotatebox[origin=c]{70}{contrast} & \rotatebox[origin=c]{70}{elastic\_trans} & \rotatebox[origin=c]{70}{pixelate} & \rotatebox[origin=c]{70}{jpeg} & Mean \\ \hline
Source& 73.0&	68.0&	39.4&	29.3&	54.1&	30.8&	28.8&	39.5&	45.8&	50.3&	29.5&	55.1&	37.2	&74.7&	41.2&	46.4\\
BN Stats Adapt & 42.1     & 40.7 & 42.7    & 27.6    & 41.9  & 29.7   & 27.9 & 34.9 & 35.0   & 41.5 & 26.5       & 30.3     & 35.7           & 32.9     & 41.2 & 35.4 \\
Pseudo-label &38.1&	36.1&	40.7&	33.2&	45.9&	38.3&	36.4&	44.0&	45.6&	52.8&	45.2&	53.5&	60.1&	58.1&	64.5&	46.2\\
TENT-continual~\cite{wang2020tent}         & \textbf{37.2}     & \textbf{35.8} & 41.7    & 37.9    & 51.2  & 48.3   & 48.5 & 58.4 & 63.7   & 71.1 & 70.4       & 82.3     & 88.0           & 88.5     & 90.4 & 60.9 \\\hline
CoTTA~(Proposed)      & 40.1&	37.7&	\textbf{39.7}&	\textbf{26.9}&	\textbf{38.0}&	\textbf{27.9}&	\textbf{26.4}&	\textbf{32.8}&	\textbf{31.8}&	\textbf{40.3}&	\textbf{24.7}&	\textbf{26.9}&	\textbf{32.5}&	\textbf{28.3}&	\textbf{33.5}&	\textbf{32.5}\\\hline
\end{tabular}\vspace{-0.3cm}\vspace{-1mm}
}
\end{table*}
\begin{table*}[h]
\centering
\caption{Semantic segmentation results (mIoU in \%) on the Cityscapes-to-ACDC online continual test-time adaptation task. We evaluate the four test conditions continually for ten times to evaluate the long-term adaptation performance. To save space, we only show the continual adaptation results in the first, fourth, seventh, and last round. Full results can be found in the supplementary material. All results are evaluated based on the Segformer-B5 architecture.}\label{tab:acdc}
\vspace{-3mm}
\scalebox{0.82}{
\tabcolsep3pt

\begin{tabular}{l|cccc|cccc|cccc|cccc|c}\hline
Time &  \multicolumn{16}{l|}{$t\xrightarrow{\hspace*{14cm}}$}\\ \hline
Round      & \multicolumn{1}{l}{1} & \multicolumn{1}{l}{} & \multicolumn{1}{l}{} & \multicolumn{1}{l|}{} & \multicolumn{1}{l}{4} & \multicolumn{1}{l}{} & \multicolumn{1}{l}{} & \multicolumn{1}{l|}{} & \multicolumn{1}{l}{7} & \multicolumn{1}{l}{} & \multicolumn{1}{l}{} & \multicolumn{1}{l|}{} & \multicolumn{1}{l}{10} & \multicolumn{1}{l}{} & \multicolumn{1}{l}{} & \multicolumn{1}{l|}{} & \multicolumn{1}{l}{All} \\ \hline
Condition          & Fog                   & Night                & rain                 & snow                  & Fog                   & Night                & rain                 & snow                  & Fog                   & Night                & rain                 & snow                  & Fog                    & Night                & rain                 & snow                  & Mean                    \\ \hline
Source          & 69.1                  & 40.3                 & 59.7                 & 57.8                  & 69.1                  & 40.3                 & 59.7                 & 57.8                  & 69.1                  & 40.3                 & 59.7                 & 57.8                  & 69.1                   & 40.3                 & 59.7                 & 57.8                  & 56.7                    \\
BN Stats Adapt & 62.3 & 38.0 & 54.6 & 53.0& 62.3 & 38.0 & 54.6 & 53.0& 62.3 & 38.0 & 54.6 & 53.0& 62.3 & 38.0 & 54.6 & 53.0 & 52.0\\
TENT-continual~\cite{wang2020tent}  & 69.0                  & 40.2                 & 60.1                 & 57.3                  & 66.5                  & 36.3                 & 58.7                 & 54.0                  & 64.2                  & 32.8                 & 55.3                 & 50.9                  & 61.8                   & 29.8                 & 51.9                 & 47.8                  & 52.3                    \\ \hline
%Weight-Averaged & 69.1                  & 40.7                 & 60.2                 & 58.3                  & 69.5                  & 41.3                 & 60.9                 & 58.6                  & 69.6                  & 41.4                 & 61.0                 & 58.7                  & 69.7                   & 41.4                 & 61.1                 & 58.7                  & 57.4                    \\
CoTTA~(Proposed)        & \textbf{70.9}& 	\textbf{41.2}& 	\textbf{62.4}& 	\textbf{59.7}& 	\textbf{70.9}& 	\textbf{41.0}& 	\textbf{62.7}& 	\textbf{59.7}& \textbf{70.9}& 	\textbf{41.0}& 	\textbf{62.8}& 	\textbf{59.7}& 	\textbf{70.8}& 	\textbf{41.0}& 	\textbf{62.8}& 	\textbf{59.7}& 	\textbf{58.6} \\ \hline
   
\end{tabular}}\vspace{-0.3cm}
\end{table*}
\noindent\textbf{Ablation study: individual components} 
The main contribution of our proposed method is to reduce the error accumulation  by using averaged pseudo-labels and random restoration. To validate our motivation, we conduct an ablation study on each of the elements of the proposed approach. As listed in Table~\ref{tab:cifar10}, by using the \textit{weight-averaged pseudo-labels} from the teacher model, the error rate is reduced from 20.7\% to 18.3\%.  This indicates that the weight-averaged predictions are indeed more accurate than the direct predictions. By using multiple augmentations to further refine the weight-averaged predictions, we are able to further improve the performance to 17.4\%. However, the performance is still deteriorating over time~(e.g. comparing to \textit{TENT-online*} for \textit{contrast}), indicating that even though the pseudo-labels are more accurate, error can still accumulate because of the inevitable wrong predictions. Finally, by using \textit{stochastic restoration} to explicitly preserve the source knowledge, the long-term predictions can be largely improved. This leads to an improved error rate of 16.2\%. The number in bracket is the standard deviation over 5 seeds.

\noindent\textbf{Gradually changing setup.}
%We agree that gradually changing is an important setup. 
In the above standard setup, corruption types change abruptly in the highest severity, we now report the results for the gradual setup. We design the sequence by gradually changing severity for the 15 corruption types: 
\begingroup
\footnotesize
${\underbrace{{\dots}2{\xrightarrow{}}1}_{\text{t-1 and before}}}{\xrightarrow[type]{\small change}}{\underbrace{1{\xrightarrow{}}2{\xrightarrow{}}3{\xrightarrow{}}4{\xrightarrow{}}5{\xrightarrow{}}4{\xrightarrow{}}3{\xrightarrow{}}2{\xrightarrow{}}1}_{\text{corruption type t, gradually changing severity}}}{\xrightarrow[type]{\small change}}\underbrace{1{\xrightarrow{}}2{\dots}}_{\text{t+1 and on}}$ ,
\endgroup
where the severity level is the lowest~(1) when corruption type changes, therefore, the  type change is gradual. The distribution shift within each type is also gradual. We create 10 randomly shuffled orders for the corruption types $t$ and then evaluate the methods using the average error rate over the ten diverse sequences. Table~\ref{gradual} shows that the proposed method outperforms competing methods, leading to an error rate of 10.4\%, compared to TENT's 30.7\%.

%\paragraph{Ablation study: confidence threshold}
%We observe that naively 

%\paragraph{Ablation study: impact of the learning rate}

\subsection{Experiments on CIFAR100-to-CIFAR100C}
To further demonstrate the effectiveness of the proposed method, we evaluate it on the more difficult CIFAR100-to-CIFAR100C task. The experimental results are summarized in Table~\ref{tab:cifar100}. We compare our method with the source-only baseline, \textit{BN stats adapt}, \textit{Pseudo-label}, as well as the \textit{TENT-continual} method. We observe that performance of the \textit{TENT-continual} model deteriorates rapidly over time on the later corruption types because of the error accumulation and forgetting. Our method yields an absolute improvement of 2.9\% error rate over \textit{BN stats adapt}, and achieves 32.5\%. More importantly, the improvement becomes larger over time, this indicates that the proposed method is able to learn from the unlabeled test images from the past streams to further improve the performance on the current test data.

\subsection{Experiments on ImageNet-to-ImageNet-C}

\begin{table}[t]
\caption{Average error of standard ImageNet-to-ImageNet-C experiments over 10 diverse corruption sequences (severity level 5).}\label{run10}
\small
\centering
\vspace{-3mm}
\scalebox{0.74}{
\tabcolsep 4pt
\begin{tabular}{c|c|c|c|c|c}
\hline
Avg. Error (\%) & Source & BN Adapt & Test Aug~[5] & TENT~[58]      & CoTTA           \\\hline
%CIFAR10-C   & 43.5   & 20.4     & 20.1   & 20.3 $\pm$ 1.2 & 16.3 $\pm$ 0.2 (0.1) \\\hline
ImageNet-C & 82.4   & 72.1     & 71.4   & 66.5 & 63.0 $\pm$ 1.8 (0.1) \\\hline

\hline
\end{tabular}}
\vspace{-3.5mm}
\end{table}

To provide a more comprehensive evaluation on the proposed method, ImageNet-to-ImageNet-C experiments are conducted over ten diverse corruption type sequences in severity level of 5. As shown in Table~\ref{run10}, CoTTA is able to continually outperform \textit{TENT} and other competing methods. The number after $\pm$ is the standard deviation over 10 diverse corruption type sequences. 

\subsection{Experiments on Cityscapes-to-ACDC}
We additionally evaluate our method on the more complex continual test-time semantic segmentation Cityscapes-to-ACDC task. The experimental results are summarized in Table~\ref{tab:acdc}. The results demonstrate that our method is also effective for semantic segmentation tasks and is robust to the different choices of architectures. Our proposed method yields an absolute improvement of 1.9\% mIoU  over the baseline, and achieves 58.6\% mIoU. It is worth mentioning that \textit{BN Stats Adapt} and \textit{TENT} do not perform well in this task and the performance deteriorates significantly over time. This is partly because both were specifically designed for networks with Batch Normalization layers, while there is only one Batch Normalization layer in Segformer and  the majority of normalization layers in transformer models are based on LayerNorm~\cite{ba2016layer}. Our method, however, does not rely on specific layers and can still be effective for this more complex task on a very different architecture. The improved performance is also largely maintained after being continually adapted for a relatively long term.

%
%\subsection{Comparison with offline methods}
%We additionally compare our methods with the existing methods on the offline CIFAR10C and CIFAR100C experiments. The comparison methods include domain adaptation based methods~(RG, UDA-SS), test-time training methods~(), and test-time adaptation methods~(TTT~\cite{sun}). 

\section{Conclusion}
In this work, we focused on the continual test-time adaptation in non-stationary environments where the target domain distribution can continually change over time. To tackle the error accumulation and catastrophic forgetting in this setup, we proposed a novel method CoTTA which comprises two parts. Firstly, we reduced the error accumulation by using weight-averaged and augmentation-averaged predictions which are often more accurate. Secondly, to preserve the knowledge from the source model, we stochastically restored a small part of the weights to the source pre-trained weights.  The proposed method can be incorporated in off-the-shelf pre-trained models without requiring any access to source data. The effectiveness of CoTTA was validated on four classification and one segmentation tasks.

\noindent
\textbf{Acknowledgement} 
The contributions of Qin Wang and Olga Fink were funded by the Swiss National Science Foundation Grant PP00P2\_176878. This work is also funded by Toyota Motor Europe via the project TRACE-Zurich.

%%%%%%%%% REFERENCES
{\small
\bibliographystyle{ieee_fullname}
\bibliography{egbib}
}

\end{document}